\documentclass{article}

    \usepackage[final]{neurips_2025}

\usepackage[utf8]{inputenc} % allow utf-8 input
\usepackage[T1]{fontenc}    % use 8-bit T1 fonts
\usepackage{hyperref}       % hyperlinks
\usepackage{url}            % simple URL typesetting
\usepackage{booktabs}       % professional-quality tables
\usepackage{amsfonts}       % blackboard math symbols
\usepackage{nicefrac}       % compact symbols for 1/2, etc.
\usepackage{microtype}      % microtypography
\usepackage{xcolor}         % colors
\usepackage{graphicx}
\usepackage{array} % newly added
\usepackage{colortbl}
\usepackage{bookmark}
\usepackage{wrapfig}
\usepackage{amsmath}
\usepackage{colortbl}
\usepackage{natbib}
\usepackage[export]{adjustbox}
\usepackage{float}

\newfloat{figtab}{htb}{fgtb}
\makeatletter
  \newcommand\figcaption{\def\@captype{figure}\caption}
  \newcommand\tabcaption{\def\@captype{table}\caption}
\makeatother

\setcitestyle{numbers,square}
\definecolor{mygray}{gray}{.9}

\title{U-REPA: Aligning Diffusion U-Nets to ViTs}

\author{%
  Yuchuan Tian$^{1}$, Hanting Chen$^{2}$, Mengyu Zheng$^{3}$, Yuchen Liang$^{4}$, Chao Xu$^{1}$, Yunhe Wang$^{2}$\\
  \small$^1$ State Key Lab of General AI, School of Intelligence Science and Technology, Peking University. \\
  \small$^2$ Huawei Noah's Ark Lab. \small$^3$ The University of Sydney. \small$^4$ School of Mathematical Sciences, Peking University. \\
  \small\texttt{tianyc@stu.pku.edu.cn, \{chenhanting, yunhe.wang\}@huawei.com}\\
  \small\texttt{xuchao@cis.pku.edu.cn}
}

\begin{document}

\maketitle

\begin{abstract}
  Representation Alignment (REPA) that aligns Diffusion Transformer (DiT) hidden-states with ViT visual encoders has proven highly effective in DiT training, demonstrating superior convergence properties, but it has not been validated on the canonical diffusion U-Net architecture that shows faster convergence compared to DiTs. However, adapting REPA to U-Net architectures presents unique challenges: (1) different block functionalities necessitate revised alignment strategies; (2) spatial-dimension inconsistencies emerge from U-Net's spatial downsampling operations; (3) space gaps between U-Net and ViT hinder the effectiveness of tokenwise alignment. To encounter these challenges, we propose \textbf{U-REPA}, a representation alignment paradigm that bridges U-Net hidden states and ViT features as follows: Firstly, we propose via observation that due to skip connection, the middle stage of U-Net is the best alignment option. Secondly, we propose upsampling of U-Net features after passing them through MLPs. Thirdly, we observe difficulty when performing tokenwise similarity alignment, and further introduces a manifold loss that regularizes the relative similarity between samples. Experiments indicate that the resulting U-REPA could achieve excellent generation quality and greatly accelerates the convergence speed. With CFG guidance interval, U-REPA could reach $FID<1.5$ in 200 epochs or 1M iterations on ImageNet 256 $\times$ 256, and needs only \textbf{half} the total epochs to perform better than REPA under \textit{sd-vae-ft-ema}. Codes: \url{https://github.com/YuchuanTian/U-REPA}

\end{abstract}

\section{Introduction}
\label{sec:intro}
Representation Alignment (REPA)~\cite{repa}, a methodology that aligns features from Diffusion Transformers (DiT)~\cite{dit} to modern visual encoders, has been demonstrated to significantly accelerate DiT training. This approach holds particular significance given the growing prominence of DiTs, which have gained mainstream adoption in diffusion models and are extensively applied across image generation~\cite{pixartalpha,sd3,flux} and video generation domains~\cite{opensora,opensoraplan,pyramidflow}. However, emerging empirical evidence suggests that U-Net~\cite{unet} architectures might present a more advantageous alternative to DiTs in certain scenarios~\cite{simplediffusion,hourglass,udit,dic}: U-Net-based models exhibit substantially faster convergence while achieving generation quality comparable to their transformer-based counterparts. This dichotomy motivates our core research inquiry - can modern Vision Transformer (ViT~\cite{vit})-based visual encoders be effectively adapted to guide diffusion U-Net training through alignment mechanisms similar to REPA, thereby potentially elevating the convergence speed ceiling of diffusion models?

\begin{figure}[!t]
  \centering
  % \vspace{-40pt}
  \includegraphics[width=0.7\textwidth]{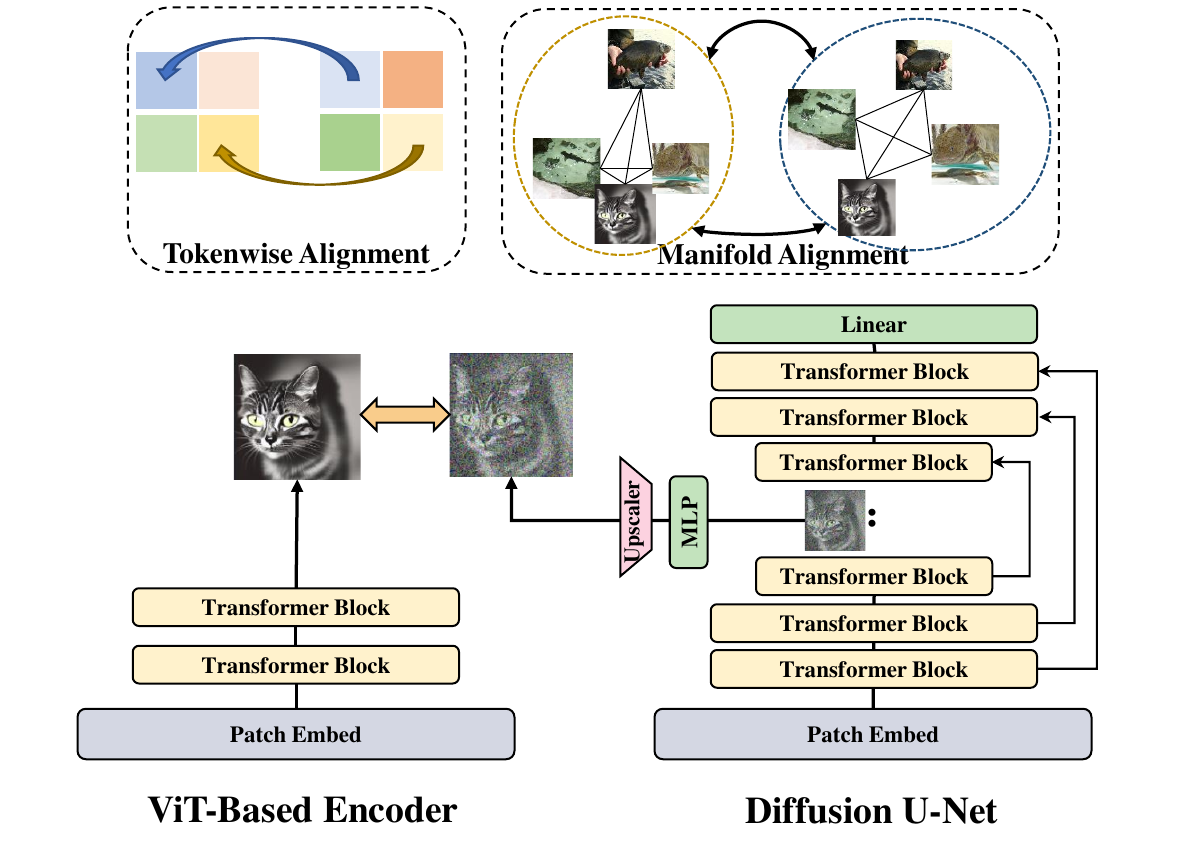}
  \vspace{-5pt}
  \caption{\textbf{The proposed U-REPA framework.} We investigated and found that semantic-rich intermediate layers are the best for representation alignment, dimension and space gaps hinders alignment efficacy. To counter these challenges, we scale-up features and propose manifold alignment.}
  \label{fig:urepa}
  \vspace{-10pt}
\end{figure}

However, establishing effective alignment between U-Net architectures and ViT-based encoders presents challenges. Unlike Diffusion Transformers (DiTs) that share structural similarities with Vision Transformers, U-Net architectures exhibit fundamentally different operational characteristics. Specifically, both DiT and ViT adopt isotropic architectures composed of uniformly stacked transformer blocks, which inherently facilitates straightforward parameter alignment between the two frameworks. In contrast, U-Net's skip connections create strong interdependencies between shallow and deep network layers by linking them together, resulting in different feature propagation dynamics. This architectural disparity renders conventional representation alignment strategies developed for DiT architectures inapplicable to U-Net frameworks. Furthermore, the progressive downsampling operations in U-Net generate feature maps with spatial dimension mismatches compared to the fixed-scale feature representations in ViT encoders, introducing additional complexity in establishing cross-architectural correspondence. In addition, features from high-stage U-Net and ViT have large space gaps, forming a barrier for cosine similarities as metrics. Forcibly using tokenwise similarity as loss is not necessarily the best option. This induces us to rethink about the optimization objective.

In order to conquer these challenges, we propose~\textbf{U-REPA}, a framework that aligns U-Net hidden states to features from ViT encoders. Firstly, Our analysis reveals that skip connections fundamentally alter the functional specialization of transformer blocks in U-Net architectures. By establishing direct dependencies between early-stage and late-stage layers, these cross-connections induce a hierarchical redistribution of semantic information, with intermediate blocks exhibiting the highest semantic density. This pattern was empirically verified through controlled ablation studies on DiT augmented with skip connections, where progressive layer-wise evaluations demonstrated peak semantic richness at median network depths.

The intermediate higher-stage layers, which contain semantically dense representations, require precise alignment with the ViT-based visual encoder. However, these critical layers undergo spatial downsampling in the U-Net architecture, necessitating explicit spatial dimension reconciliation between U-Net features and ViT features during representation alignment. Through empirical exploration of various resolution-matching strategies, we identified an optimal solution: performing linear transformation via MLP on U-Net features prior to upsampling operations, which achieves superior alignment performance compared to alternative approaches.

Further analysis revealed a fundamental incompatibility of measuring cosine similarity between the feature spaces of U-Net and ViT encoders. Enforcing strict token-wise similarity constraints proves excessively rigid due to inherent architectural discrepancies. To address this, we introduce a manifold loss that implements soft alignment through relational regularization. This loss operates on the relative geometric relationships between samples rather than imposing direct feature correspondence, thereby accommodating cross-architectural variations. Comprehensive experiments demonstrate that our proposed U-REPA framework effectively bridges the U-Net-ViT alignment gap while preserving the distinct advantages of both architectures.

Our contributions are as follows:

\begin{enumerate}
  \item We identify U-Net's representation alignment to be a challenge due to different block functionalities, spatial-dimension inconsistencies, and larger feature space gaps.
  \item We evaluate the contribution of downsampling and skips in U-Net and demonstrate U-Net's potential advantage over DiTs.
  \item We propose U-REPA, a framework that evaluates layers, investigates the best scale-up policy, and introduces manifold-space loss in aid of alignment.
  \item We conduct experiments and verify the effectiveness of our U-REPA framework in terms of fast convergence. Specifically, U-REPA reaches $FID<1.5$ in just 200 epochs on ImageNet 256 $\times$ 256; and it reaches 1.41 FID, with only half the epochs of REPA under the same training setting.
\end{enumerate}

\section{Related Work}
\label{sec:relatedwork}
\noindent\textbf{The development of diffusion architectures.} The conventional diffusion works~\cite{ddpm,ddim,songunet,dhariwal} leverages a U-Net~\cite{unet} architecture, whose basic block is a concatenation of convolution layers and self-attention. More recent architectural innovations in diffusion models have witnessed a paradigm shift from conventional U-Net frameworks toward transformer-based architectures. The emergence of Diffusion Transformers~\cite{uvit,dit} demonstrates their competitive performance despite abandoning the inductive biases inherent in U-Net designs. U-ViT~\cite{uvit} represents an intermediate architecture that preserves U-Net's hierarchical structure but replaces convolutional blocks with transformer layers, notably omitting the traditional downsampling operations. Subsequent developments have further streamlined the architecture: DiT~\cite{dit} adopts a pure transformer backbone with isotropic scaling, while SiT~\cite{sit} integrates the transformer architecture into the RectifiedFlow framework. Some other works either improve the micro-designs~\cite{visionllama,fit}, or focuses on architectural efficiency~\cite{pixartsigma,sana,lit}.

In contrary to these DiT works, some works still sticks to U-Net architectures and offer valuable rethinking on this conventional architectural preference: in pixel-space image generation, works including SimpleDiffusion~\cite{simplediffusion} and HourglassDiT~\cite{hourglass} still sticks to U-Net; in  with its variants like U-DiT~\cite{udit}, Playground v3~\cite{playgroundv3}, and DiC~\cite{dic} extending its success to latent-space diffusion through simple Conv3$\times$3 designs. While these implementations empirically validate U-Net's accelerated convergence and stable training dynamics compared to transformer-based alternatives, current research predominantly focuses on proposing architectural modifications rather than uncovering the reasons of U-Net's superior diffusion performance.

\noindent\textbf{Techniques for better DiT performance.} Building upon the success of self-supervised learning~\cite{mae}, MDT~\cite{mdt} and MaskDiT~\cite{maskdit} pioneer masked image modeling in diffusion frameworks by adaptively masking a good proportion of input patches during training. Other than the masking strategy, a bunch of diffusion works refer to higher-level semantic guidance from off-the-shelf pretrained models that significantly improves generation quality. REPA~\cite{repa} establishes feature alignment between ViT-based encoder embeddings and diffusion latent spaces through contrastive learning. LightningDiT~\cite{vavae} innovates through an improved VAE distilled from MAE~\cite{mae} and DINOv2~\cite{dinov2}. Ma et al.~\cite{diffcot} introduces CLIP~\cite{clip} and DINO~\cite{dino} to verify inference-time scaling of diffusion models. These methods demonstrate that a higher-level semantic-rich feature-map from pretrained vision encoders is helpful to diffusion-based generation.

\section{Method}
\label{sec:method}
\subsection{Preliminaries: REPA for DiT}

Representation Alignment (REPA)~\cite{repa} distills Diffusion Transformers with semantic features from off-the-shelf ViT-based vision encoders (e.g. DINOv2~\cite{dinov2}, CLIP~\cite{clip}, MAE~\cite{mae}, et cetera). Given a ViT-based vision encoder \( f \) and clean image \( \mathbf{x}_\ast \), let \( \mathbf{y}_\ast = f(\mathbf{x}_\ast) \in \mathbb{R}^{N \times D} \) denote its patch embeddings, where \( N \) and \( D \) represent the number of patches and embedding dimension, respectively. REPA establishes feature alignment between \( \mathbf{y}_\ast \) and the projected diffusion encoder outputs \( h_\phi(\mathbf{h}_t) \), where \( \mathbf{h}_t = f_\theta(\mathbf{z}_t) \) is the latent representation from the diffusion transformer at timestep \( t \), and \( h_\phi \) is a trainable multilayer perceptron (MLP).  

The alignment is enforced by maximizing token-wise feature similarities, \textit{i.e.} the similarity of a token from DiT hidden-state with its corresponding counterpart in the ViT encoder feature:  
\begin{equation}
  \mathcal{L}_{\text{REPA}}(\theta, \phi) := -\mathbb{E}_{\mathbf{x}_\ast, {\epsilon}, t} \left[ \frac{1}{N}\sum_{n=1}^{N} \mathrm{sim}\left(\mathbf{y}_\ast^{[n]}, h_\phi(\mathbf{h}_t^{[n]})\right) \right],
  \label{eq:repa_loss}
\end{equation}  
where \( \mathrm{sim}(\cdot,\cdot) \) denotes a similarity metric (e.g., cosine similarity). Typically, \( \mathbf{z}_t \) is adopted as the output from early layers (the original work adopts layer index 8) in DiT for better alignment. This alignment term is combined with the basic flow-based diffusion objective (\textit{i.e.} SiT~\cite{sit}) through a tunable coefficient \( \lambda > 0 \), and the final loss for diffusion model training is formulated as follows:

\begin{equation}
  \mathcal{L} := \mathcal{L}_{\text{velocity}} + \lambda\mathcal{L}_{\text{REPA}}.
  \label{eq:combined_loss}
\end{equation}

\subsection{Evaluating the Potential of U-Net}

In diffusion models, U-Net and isotropic architectures (e.g., DiT) exhibit distinct design philosophies. While DiT achieves state-of-the-art results through scalability and integration with advanced techniques, U-Net-based methods emphasize faster convergence~\cite{udit}. To dissect U-Net's efficacy, we isolate its two core components: skip connections and downsampling.

\begin{enumerate}
    \item \textbf{Skip Connections:} Provide shortcuts between encoder and decoder layers, theoretically aiding gradient flow and feature reuse.
    \item \textbf{Downsampling:} Reduces spatial resolution (typically by a scale factor of 2 at each stage) to enable hierarchical, multi-scale feature learning. Critically, downsampling is always paired with skip connections to mitigate information loss.
\end{enumerate}

\begin{figure*}[t]
  \centering
  \vspace{-30pt}
  \begin{minipage}{0.49\textwidth}
    \centering
    \includegraphics[width=\linewidth]{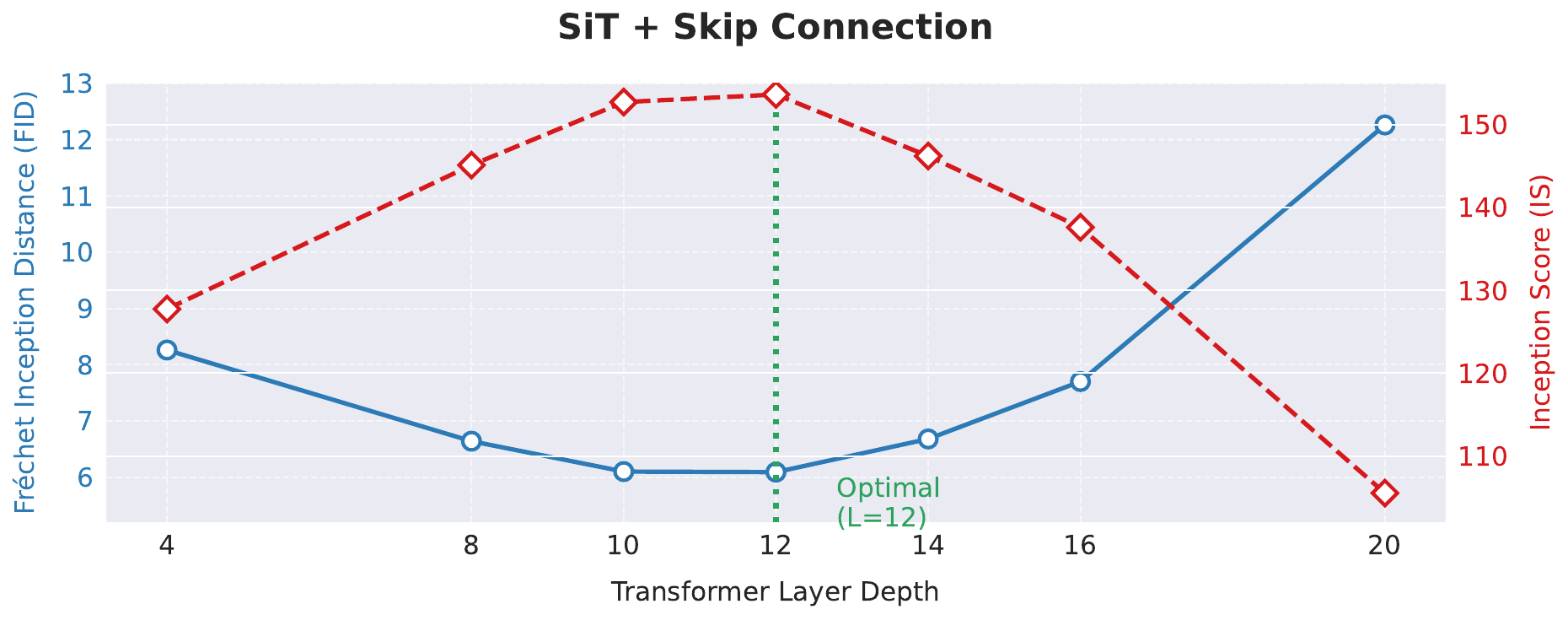}
    % \subcaption{Equal Width Subfigure 1}
    \label{fig:equal1}
  \end{minipage}
  \hfill
  \begin{minipage}{0.49\textwidth}
    \centering
    \includegraphics[width=\linewidth]{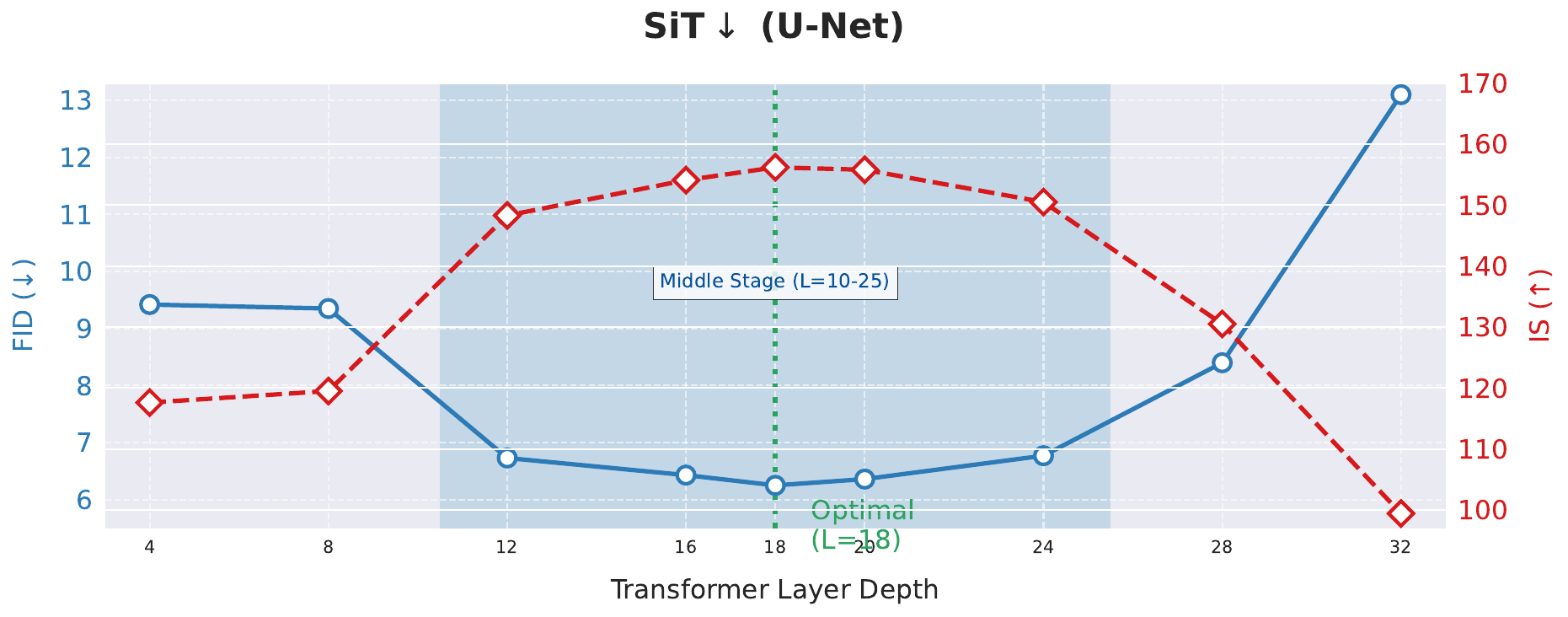}
    % \subcaption{Equal Width Subfigure 2}
    \label{fig:equal2}
  \end{minipage}
  \vspace{-5pt}
  \caption{\textbf{Investigating alignment  with respect to encoder depths on diffusion models with skip connections.} \textbf{Left:} SiT with skip connections. Due to the change of block functionalities due to newly established skip dependencies, the most optimal encoder depth is shifted towards the middle of the model. \textbf{Right:} SiT$\downarrow$, the U-Net-based SiT model. Shadowed region represents higher U-Net stage. The plot infers that stage transitions (downsampling\& upsampling in U-Net) bring large block functionality gaps. Alignment within higher U-Net stage is thus necessary for alignment performance.}
  \label{fig:advanced}
  \vspace{-10pt}
\end{figure*}

\noindent\textbf{Toy experiments on U-Net components.} On top of DiT, we perform toy experiments that reveal the contribution of components mentioned above.

\begin{table}[htbp]
    \centering
    \setlength{\belowcaptionskip}{0cm}   
  \begin{tabular}{lccc}
    \toprule
    \multicolumn{4}{l}{\bf{ImageNet} 256$\times$256, DiT 400K, cfg=1} \\
    \toprule
    Model & FLOPs (G) & FID$\downarrow$ & IS$\uparrow$ \\
    \midrule
    \textbf{DiT-XL/2} & 118.6 & 19.47 & - \\
    \textbf{DiT-XL/2$^*$} & 118.6 & 20.05 & 66.74 \\
    \textbf{+ Skip Connections} & 114.1 & 19.86 & 67.29 \\
    \textbf{+ Downsampling} & 108.8 & 13.78 & 88.93 \\
    \midrule
    \rowcolor{gray!20}\textbf{DiT$\downarrow$-XL/2 (+Tricks)} & 108.8 & \textbf{11.02} & \textbf{100.35} \\
    \bottomrule
    \end{tabular}
    \vspace{5pt}
    \caption{\textbf{Evaluating the contribution of U-Net components in terms of fast convergence.} Experiments are conducted using hyperparameters from~\cite{dit} for 400K iterations. Model depth is changed when a modification is made such that the overall FLOPs is kept almost the same with DiT.}
    \label{tab:method_conditioning}
    \vspace{-5pt}
  \end{table}

This suggests that U-Net's fast-convergence advantages primarily stem from multi-scale hierarchical modeling via downsampling, not skip connections. Downsampling compresses features into compact, semantically rich representations, accelerating learning while maintaining information flow through skip-augmented decoder layers. However, skip connection is not useless as it compensates for the information loss due to downsampling.

Building on this insight, we propose DiT$\downarrow$ (and SiT$\downarrow$ for the flow-based version, following the naming convention of~\cite{sit}) by adding tricks of RoPE~\cite{rope} and SwiGLU following previous work~\cite{visionllama,udit}

\subsection{Aligning U-Net to ViT Encoders}
Since U-Net has good potentials to achieve excellent generation, we are motivated to investigate whether REPA could also work on U-Net. We are first focused on the block functionality pattern and investigates the most optimal position for alignment; then we are interested in feature size alignment problems; lastly, we are dedicated to merging space gaps between features from U-Net and ViT, respectively.

\noindent\textbf{Position for alignment.} Regarding the concern that block functionalities differ between U-Net and DiT, the comparison of prior studies~\cite{diffpruning,deltadit} demonstrates divergent hierarchical specialization: U-Net architectures typically employ mid-network layers for high-level semantic synthesis while reserving shallow layers for low-level image refinement, whereas DiTs exhibit a totally different pattern - early layers primarily govern semantic-rich outline formation with deeper layers handling detailed image refinement. 

These previous findings find empirical support in REPA's experimental findings, where representation alignment proves most effective when applied to initial transformer blocks. This phenomenon stems from DiT's early layers encoding semantic-rich representations that align well with the semantically dense outputs of ViT-based visual encoders, enabling meaningful guidance. Unlike the straightforward DiT architecture, the inherent skip connections in U-Net architectures induce fundamentally distinct block functionality compared to Diffusion Transformers. While all blocks in ViT or DiT maintain homogeneous computational roles, following a continuous flow of transition from input to output, U-Net's cross-layer shortcuts establish direct dependencies between shallow and deep layers, fundamentally altering feature-map evolution patterns. As shown in Fig.~\ref{fig:advanced} (R), DiTs with skips indicates median layer is the best for representation alignment. The same pattern goes for U-Net (Fig.~\ref{fig:advanced} (L)) despite the downsampling stage.

\noindent\textbf{Feature size alignment.} The implementation of alignment between Diffusion U-Net's median stage and ViT encounters a critical spatial resolution dilemma stemming from architectural disparities. While our analysis identifies mid-network U-Net features as optimal semantic carriers, their spatial dimensions drastically differ from ViT's full-resolution token sequence. This dimensional mismatch obstructs REPA's token-wise similarity computation, which requires strict cardinality matching between compared features.

In order to align two feature-maps (\textit{i.e.} from U-Net and ViT encoder, respectively), from the macro level we advocate for upscaling the smaller-sized U-Net features rather than downscaling the larger visual encoder features. This design principle stems from the critical observation that compressing ViT's high-resolution features to match U-Net's reduced dimensions inevitably discards fine-grained visual information, thereby degrading alignment effectiveness. Preserving ViT's native resolution while expanding U-Net's bottleneck features proves essential for maintaining semantic fidelity.

At the implementation level, we empirically evaluated various upscaling strategies for U-Net features:
\begin{enumerate}
  \item \textbf{Upscale first and then MLP:} feature upsampling is performed before passing the feature into the MLP.
  \item \textbf{Upscale within MLP:} the MLP also acts as a feature upsampler that receives a low-resolution input and outputs a high-resolution one via linear mapping and pixel un-shuffling.
  \item \textbf{MLP first and then upscale:} the feature from higher-stage U-Net is first passed through MLP and then upsampled. 
\end{enumerate}

Among the three options, we found ``MLP first and then upscale'' is the best both in terms of performance and efficiency (minimum FLOPs cost), which will be discussed in the Ablation Study in Sec.~\ref{sec:experiments}.

\noindent\textbf{Manifold space alignment.} Though we select the most suitable U-Net feature for alignment and keep dimensions between U-Net and ViT features aligned, challenges remain in feature space compatibility. First, compared to the structural congruence between DiT and ViT encoders, the architectural discrepancy of U-Net (with its skip connections and hierarchical downsampling) creates a more pronounced feature distribution gap between U-Net hidden states and visual encoder outputs. Second, the dimensional transformation required for alignment inevitably modifies U-Net's native feature space characteristics. 

Some recent works on Diffusion U-Net~\cite{unified,freeu} reveals that higher-stage U-Net features are low-frequency subspaces that discards higher frequency compoenents, including noises. Gaps are inevitable when evaluating the cosine similarities of detail-rich, high-frequency-rich vectors and flat, low-frequency-dominated vectors. In this sense, strict token-level alignment constraints like the original REPA loss prove suboptimal under these conditions, as they assume implicit feature space homogeneity between aligned modalities.

\begin{figure}[!t]
  \centering
  % \vspace{-30pt}
  \includegraphics[width=0.45\textwidth]{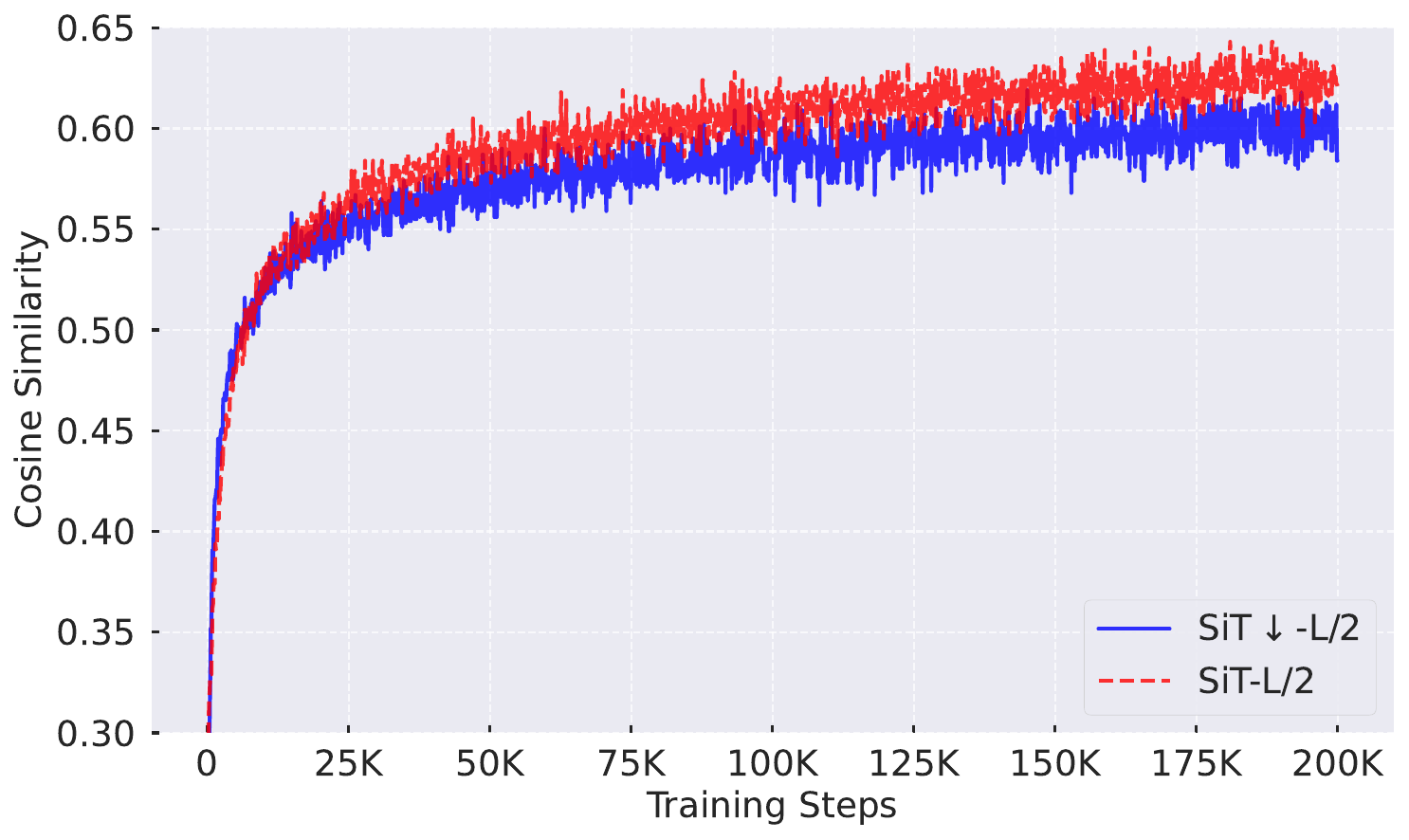}
  \vspace{-5pt}
  \caption{\textbf{The convergence of average tokenwise similarities.} While SiT-L/2 could achieve better tokenwise similarities, SiT$\downarrow$ converges at a lower similarity value, indicating difficulties of feature alignment.}
  \label{fig:cossim}
  \vspace{-10pt}
\end{figure}

As is shown in Fig.~\ref{fig:cossim}, we conducted continuous measurements of token-wise cosine similarity against ViT features during training. The two models that we compare are SiT-L/2 of isotropic, standard transformer architecture and SiT$\downarrow$-L/2 of U-Net architecture. Our experiments revealed a characteristic learning trajectory: while U-Net achieves slightly faster similarity improvement in early training phases - thanks to skip connection that helps convergence - its progress stagnates beyond this point, ultimately plateauing at 0.60 - notably inferior to DiT's sustained growth reaching around 0.63 similarity. The similarity gap between SiT and SiT$\downarrow$ This phenomenon suggests that naively aligning U-Net with ViT encoders through angular similarity metrics alone encounters inherent limitations due to architectural incompatibilities.

Rather than strict token-wise regularization, we resort to looser objectives that does not require rigid augular alignment. Inspired by manifold knowledge distillation~\cite{manifoldKD}, we hold that aligning similarities between samples from the same feature space could be a promising solution. Hence, we define Manifold Loss $\mathcal{L}_{\text{ML}}$ as

\begin{equation}
  \mathcal{L}_{\text{ML}}(\theta, \phi) := -\mathbb{E}_{\mathbf{x}_\ast, {\epsilon}, t, i, j} \left[ \mathrm{d}(\mathbf{y}_\ast, h_\phi(\mathbf{h}_t)) \right],
  \label{eq:mani_loss}
\end{equation}  
where
\begin{equation}
  \mathrm{d} := \| \mathrm{sim}\left(\mathbf{y}_\ast^{[i]}, \mathbf{y}_\ast^{[j]}\right) - \mathrm{sim}\left( h_\phi(\mathbf{h}_t^{[i]}), h_\phi(\mathbf{h}_t^{[j]})\right) \|_F^2.
  \label{eq:mani_dist}
\end{equation}  

In the formula, cosine similarity is adopted as the similarity metric, and $F$ represents Frobenius Norm of matrices. By introducing affine hyperparameter $w$, the overall optimization target is then formulated as

\begin{equation}
  \mathcal{L} := \mathcal{L}_{\text{velocity}} + \lambda\left(\mathcal{L}_{\text{REPA}}+w \mathcal{L}_{\text{ML}}\right).
  \label{eq:all_loss}
\end{equation}

\subsection{Other Improvements}

We also propose and evaluate some other improvements. Due to page limits, the proposed methods and corresponding ablations are enclosed in the Appendix.

\section{Experiments}
\label{sec:experiments}

\subsection{Experiment Setup}

\noindent\textbf{Experiment settings.} Our implementation completely adheres to the training protocol established in REPA~\cite{repa}. Following the architectural configuration of latent diffusion models~\cite{sd}, we employ the identical VAE variant (\textit{sd-vae-ft-ema}) and adopt the AdamW optimizer. To ensure fair comparison, we maintain identical hyperparameter settings across all experiments: a global batch size of 256, fixed learning rate of $1e-4$, and disabled weight decay (set to 0). ($\beta_1, \beta_2$) is set as (0.9, 0.999). All experiments are conducted on the ImageNet 2012 benchmark~\cite{imagenet} under a controlled environment with a fixed random seed (global seed=0). 8 NVIDIA A100 GPUs are used for main experiments.

For main experiments (Tab.~\ref{tab:cfg}), we apply guidance interval~\cite{guidanceinterval} $[0,0.7]$ and SDE sampling according to the convention of REPA~\cite{repa} for fair comparison. We select smaller cfg of 1.65, because we found it is better for our architecture, different from SiT. For all ablation experiments, we train models for 100K iterations, which is sufficient to show the trend of model performance; sampling is conducted with the default setting of the official REPA codebase, \textit{i.e.} $cfg=1.8$ in ODE and guidance interval $[0,0.7]$.

\noindent\textbf{Model settings.} By aligning channel dimensions and FLOPs with standard Diffusion Transformers (DiTs or SiTs), our U-REPA-compatible variants maintain architectural parity while introducing critical adaptations for U-Net principles. The base model (SiT$\downarrow$-B) employs a stage arrangement of [5,5,5], achieving 199.7G FLOPs. Scaling to larger models, we have the L variant (686.6M) and XL variant (954.4M params) that increases channel width (1024 vs. 1152 in base) through increased stage-wise block allocation ([9,14,9] vs. [10,16,10]). Notably, when FLOPs are aligned, SiT$\downarrow$ models usually have more parameters than SiTs due to increased depth.

\begin{table*}[htbp]
	\centering
	% \footnotesize
	\setlength{\belowcaptionskip}{0cm}
	\begin{tabular}{ccccccc}
		\toprule
		Model & Params (M) & FLOPs (G) & Patch Size & Channel & \# Heads & Blocks in Stages \\
		\midrule
		\textbf{SiT$\downarrow$-B} & 199.7 & 24.1 & 2 & 768 & 12 & [5,5,5] \\
		\textbf{SiT$\downarrow$-L} & 686.6 & 79.3 & 2 & 1024 & 16 & [9,14,9] \\
		\textbf{SiT$\downarrow$-XL} & 954.4 & 109.3 & 2 & 1152 & 16 & [10,16,10] \\
		\bottomrule
	\end{tabular}
	\vspace{-5pt}
	\caption{\textbf{Configurations of SiT$\downarrow$ architecture at different model sizes.} The proposed SiT$\downarrow$ in U-Net architectures are aligned to DiTs in terms of FLOPs and channel dimension.}
	\label{tab:configurations}
  \vspace{-10pt}
\end{table*}

\subsection{The Advantage of U-REPA}

\noindent\textbf{Comparing SiT$\downarrow$ with SiT at different scales.} We evaluate our U-REPA alignment method on ImageNet 256 under a generation setting with $cfg=1$ (REPA framework without classifier-free guidance). As shown in Table~\ref{tab:nocfg}, our approach consistently improves generation quality while significantly reducing computational costs across model scales. For the base-size SiT-B/2 variant, integrating U-REPA achieves a 39.3\% improvement in FID (from 24.4 to 15.3) with comparable FLOPs (24.1G vs. 23.0G) and identical training iterations (400K), demonstrating that feature alignment enhances parameter efficiency without additional training overhead. The acceleration effect becomes more pronounced in larger models: for SiT-L/2, U-REPA reduces required iterations by 42.9\% (700K→400K) while simultaneously lowering FLOPs (79.3G vs. 80.8G) and achieving a 30.9\% FID improvement (8.4→5.8). Most notably, the XL-scale variant with U-REPA (\textit{cf.} Fig.~\ref{fig:convergence} for FIDs vs. Training iters) attains state-of-the-art FID (5.4) using \textbf{90\%} fewer iterations (400K vs. 4M) and fewer FLOPs (108.8G vs. 118.6G) compared to the baseline, proving our method's fast convergence.

We also demonstrate the advantage of the proposed U-REPA framework when measuring by parameters (rather than computation FLOPs), as shown in Fig.~\ref{fig:paramscale}. Though U-Net brings extra parameters when FLOPs are aligned with DiTs, the advantage of SiT$\downarrow$+U-REPA is obvious as depicted in the Parameter versus FID plot.

\begin{figtab}[t]
  % \vspace{-30pt}
  \begin{minipage}[t]{0.35\linewidth}
    \vspace{-10pt}
  \centering
  \includegraphics[width=\textwidth]{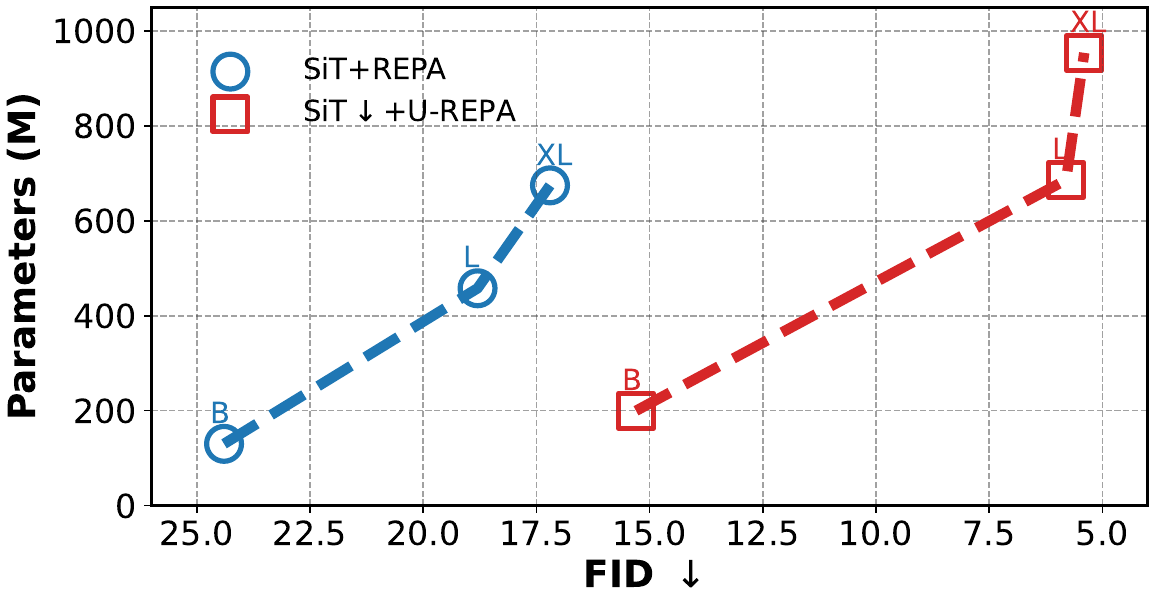}
  \figcaption{\textbf{Comparing SiT$\downarrow$+U-REPA against SiT+REPA in terms of parameter scalability.} While the U-Net architecture makes the diffusion model parameter-rich compared with same-FLOPs Diffusion Transformers, SiT$\downarrow$ models still outcompetes SiTs by large margins in terms of parameters.}
  \label{fig:paramscale}
\vspace{10pt}
    \centering
    \includegraphics[width=\textwidth]{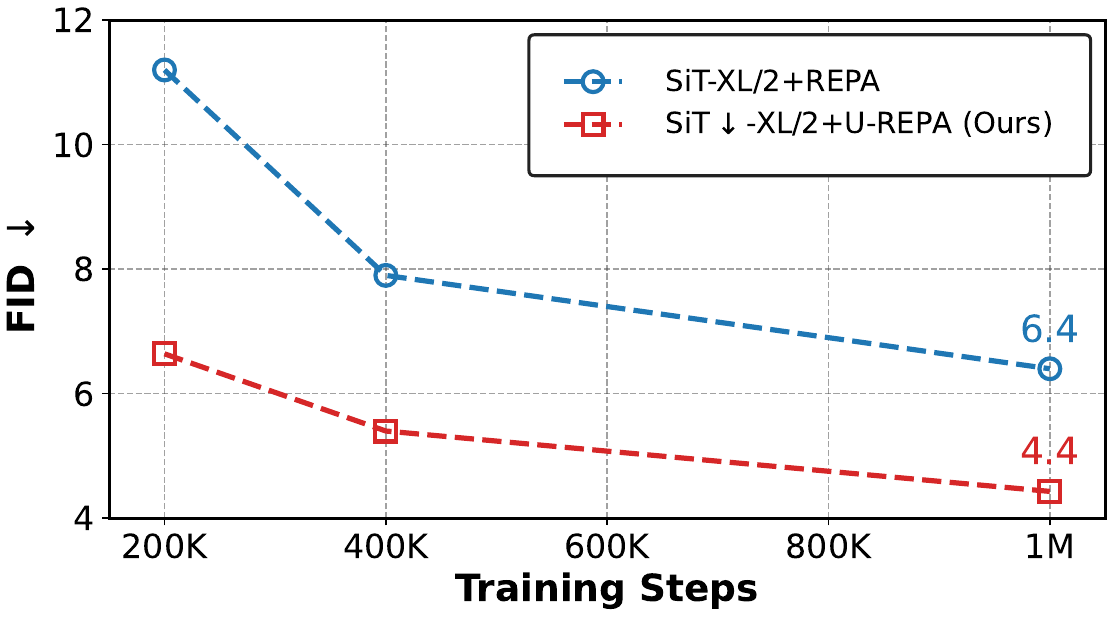}
    \figcaption{\textbf{Comparing SiT$\downarrow$+U-REPA against SiT+REPA in terms of convergence speed.} SiT$\downarrow$-XL/2 convergences much faster than SiT-XL/2 with the help of U-REPA.}
    \label{fig:convergence}
  \end{minipage}
  \quad
  \begin{minipage}[t]{0.62\linewidth}
    \vspace{-10pt}
    \centering
    \begin{tabular}{lccc}
      \toprule
      \multicolumn{4}{l}{\bf{ImageNet} 256$\times$256, w/o cfg} \\
      \toprule
      Model & FLOPs (G) & Iter. & FID$\downarrow$ \\
      \midrule
      \textbf{SiT-B/2+REPA} & 23.0 & 400K & 24.4 \\
      \rowcolor{gray!20}\textbf{SiT$\downarrow$-B/2+U-REPA} & 24.1 & 400K & \textbf{15.3} \\
      \midrule
      \textbf{SiT-L/2+REPA} & 80.8 & 700K & 8.4 \\
      \rowcolor{gray!20}\textbf{SiT$\downarrow$-L/2+U-REPA} & 79.3 & 400K & \textbf{5.8} \\
      \midrule
      \textbf{SiT-XL/2+REPA} & 118.6 & 4M & 5.9 \\
      \rowcolor{gray!20}\textbf{SiT$\downarrow$-XL/2+U-REPA} & 108.8 & 400K & \textbf{5.4} \\
      \bottomrule
      \end{tabular}
      % }
      % \vspace{-10pt}
    
      \tabcaption{\textbf{Comparing U-REPA against REPA across various model sizes without classifier-free guidance.} U-Nets equipped with U-REPA show excellent capabilities. Notably, U-REPA achieves 10$\times$ faster convergence compared with REPA in terms of performance w/o CFG.}
      \label{tab:nocfg}
      \vspace{10pt}
      \begin{tabular}{lccc c}
        \toprule
        \multicolumn{5}{l}{\bf{ImageNet} 256$\times$256, w/ cfg} \\
        \toprule
        Model & Dep. & Feat. Dim. & FID$\downarrow$ & IS$\uparrow$ \\
        \midrule
        \textbf{SiT$\downarrow$-XL/2} & 4 & 16 & 9.42 & 117.6 \\
        \rowcolor{gray!20}\textbf{SiT$\downarrow$-XL/2} (REPA) & 8 & 16 & 9.35 & 119.5 \\
        \textbf{SiT$\downarrow$-XL/2} & 12 & 8 & 6.73 & 148.3 \\
        \textbf{SiT$\downarrow$-XL/2} & 16 & 8 & 6.43 & 154.1 \\
        \rowcolor{gray!20}\textbf{SiT$\downarrow$-XL/2} & \textbf{18} & \textbf{8} & \textbf{6.25} & \textbf{156.2} \\
        \textbf{SiT$\downarrow$-XL/2} & 20 & 8 & 6.36 & 155.8 \\
        \textbf{SiT$\downarrow$-XL/2} & 24 & 8 & 6.77 & 150.5 \\
        \textbf{SiT$\downarrow$-XL/2} & 28 & 16 & 8.40 & 130.5 \\
        \textbf{SiT$\downarrow$-XL/2} & 32 & 16 & 13.10 & 99.4 \\
        \bottomrule
      \end{tabular}
      % }
      \caption{\textbf{Ablations on encoder depths for alignment in SiT$\downarrow$.} Feat. Dim. stands for the spatial height\& width at the certain layer. Compared with the default REPA setting, aligning at the centering layer (higher stage in U-Net) performs much better.}
      \label{tab:layer}
\end{minipage}
\vspace{-10pt}
\end{figtab}

\noindent\textbf{Convergence performance.} We also compare our method with previous State-of-the-Arts, as shown in Tab.~\ref{tab:cfg}. Our proposed SiT$\downarrow$+U-REPA achieves a competitive FID of 1.48 with only 200 training epochs, significantly outperforming existing methods in training efficiency. Notably, while state-of-the-art masked diffusion transformers like MDTv2-XL/2 require 1,080 epochs to reach 1.58 FID, our method attains better performance (1.48) with 80\% fewer iterations. Even compared to the SOTA SiT-XL/2 + REPA baseline (800 epochs for 1.42 FID), our approach uses only 1/2 of the training epochs (400) while achieving better generation quality (1.41 FID). The results demonstrate that the proposed U-REPA establishs a new efficiency frontier for diffusion models.

\subsection{Ablation Studies}

\noindent\textbf{Encoder depths.} The ablation study on encoder layer depths for feature alignment (Tab.~\ref{tab:layer}) coincides with the pattern of DiT with skip connections, as we analyzed in Sec.~\ref{sec:experiments}: despite progressive downsampling operations that reduce spatial resolution, the centermost layers exhibit optimal alignment efficacy. For the SiT$\downarrow$-XL/2 model, aligning features at layer 18 (midway through the 36-layer architecture) achieves peak performance with 6.25 FID and 156.2 IS, outperforming both shallower and deeper alignment points. This phenomenon persists even as the spatial dimension (Feat. Dim.) halves from 16$\times$16 to 8$\times$8 in the intermediate stage, indicating that semantic richness—not spatial resolution—dominates alignment quality. Performance degradation occurs when alignment takes place at shallower or deeper stages, even though the feature size is kept the same with DINO in these stages.

\begin{table}[!t]
  % \vspace{-30pt}
  \begin{minipage}[t]{0.5\linewidth}
    \vspace{40pt}
    \centering
    \setlength{\tabcolsep}{4pt}
  \begin{tabular}{lcc}
        \toprule
        \multicolumn{3}{l}{\bf{ImageNet} 256$\times$256, w/ cfg} \\\midrule
        { Model} & Epochs  &  { FID$\downarrow$} \\
        \arrayrulecolor{black}\midrule
        
        \multicolumn{3}{l}{\emph{Pixel diffusion}\vspace{0.02in}} \\
         ADM-U~\cite{dhariwal}      & 400  & 3.94 \\
         VDM$++$~\cite{vdmpp}   & 560  & 2.40 \\
         Simple diffusion~\cite{simplediffusion} & 800  & 2.77 \\
        \arrayrulecolor{black!30}\cmidrule(lr){1-3}
        \emph{Latent Diffusion Transformer} \\
         U-ViT-H/2~\cite{uvit}  & 240  & 2.29 \\ 
         DiffiT~\cite{diffit}  & -    & 1.73 \\
         DiT-XL/2~\cite{dit}   & 1400 & 2.27 \\
         SiT-XL/2~\cite{sit}   & 1400 & 2.06 \\
        %  \arrayrulecolor{black}\midrule
        \arrayrulecolor{black!30}\cmidrule(lr){1-3}
    
         \emph{Masked Diffusion Transformer} \\
         MaskDiT~\cite{maskdit} & 1600 & 2.28 \\ 
         MDTv2-XL/2~\cite{mdt}& 1080 & 1.58 \\
        % \arrayrulecolor{black}\midrule
    
        \arrayrulecolor{black!30}\cmidrule(lr){1-3}
        \emph{Representation Alignment} \\
         SiT-XL/2 + REPA~\cite{repa}       & 800  & \underline{1.42} \\
         \rowcolor{gray!20}{SiT$\downarrow$-XL/2 + U-REPA (Ours)} & {200}  & {1.48} \\
         \rowcolor{gray!20}\textbf{SiT$\downarrow$-XL/2 + U-REPA (Ours)} & \textbf{400}  & \textbf{1.41} \\
        \arrayrulecolor{black}\bottomrule
      \end{tabular}
      \vspace{5pt}
    \tabcaption{\textbf{Comparing U-REPA against State-of-the-Art baselines with classifier-free guidance.} U-REPA could reach $FID<1.5$ in merely 200 epochs and $FID=1.41$ in 400 epochs; The proposed method converge $2\times$ faster while achieving lower FID.}
    \label{tab:cfg}
    \end{minipage}\quad
    \begin{minipage}[t]{0.47\linewidth}
      \vspace{-5pt}
  \centering 
  % \resizebox{\textwidth}{!}{%
\begin{tabular}{lcc}
  \toprule
  \multicolumn{3}{l}{\bf{ImageNet} 256$\times$256, w/ cfg} \\
  \toprule
  Alignment Choices & FID$\downarrow$ & IS$\uparrow$ \\
  \midrule
  \textbf{U-Net $ || $ DINOv2$\downarrow_2$} & 5.99 & 158.8 \\
  \rowcolor{gray!20}\textbf{U-Net$\uparrow_2 || $ DINOv2} & \textbf{5.72} & \textbf{161.6} \\
  \bottomrule
  \end{tabular}
  % }
  \vspace{2pt}
  \caption{\textbf{Alignment dimension choices.} Upsampling higher-stage U-Net features in alignment with ViT performs better due to less information loss.}
  \label{tab:dimension}

  \begin{tabular}{lcc}
    \toprule
    \multicolumn{3}{l}{\bf{ImageNet} 256$\times$256, w/ cfg} \\
    \toprule
    Alignment & FID$\downarrow$ & IS$\uparrow$ \\
    \midrule
    \textbf{Upscale before MLP} & 5.84 & 158.5 \\
    \textbf{Upscale in MLP} & 6.36 & 153.4 \\
    \rowcolor{gray!20}\textbf{Upscale after MLP} & \textbf{5.72} & \textbf{161.6} \\
    \bottomrule
    \end{tabular}
    % }
    \vspace{2pt}
    \caption{\textbf{Feature-map upscale choices.} Among the three options, Upscaling after passing through MLP performs best; and it has lower cost as the small-sized feature map is passed through MLP.}
    \label{tab:upscale}

    \begin{tabular}{lccc}
      \toprule
      \multicolumn{4}{l}{\bf{ImageNet} 256$\times$256, w/ cfg} \\
      \toprule
      Model & $w$ & FID$\downarrow$ & IS$\uparrow$ \\
      \midrule
      \textbf{SiT$\downarrow$-XL/2+U-REPA} & 0 & 6.25 & 156.2 \\
      \midrule
      \textbf{SiT$\downarrow$-XL/2+U-REPA} & 2 & 5.81 & 160.8 \\
      \rowcolor{gray!20}\textbf{SiT$\downarrow$-XL/2+U-REPA} & \textbf{3} & \textbf{5.72} & \textbf{161.6} \\
      \textbf{SiT$\downarrow$-XL/2+U-REPA} & 4 & 5.79 & 160.6 \\
      \bottomrule
      \end{tabular}
      % }
      \caption{\textbf{Adjusting weight $w$ in Eq.~\ref{eq:all_loss}.} Manifold loss boosts U-Net's alignment performance. The most optimal result is taken at $w=3$.}
      \label{tab:weight}
\end{minipage}
  \vspace{-20pt}
\end{table}

\noindent\textbf{Alignment dimension choices.} The comparative results in Table~\ref{tab:dimension} reveal that upsampling U-Net's higher-stage features ($\uparrow_2$) to match DINOv2's native resolution achieves superior performance (5.72 FID, 161.6 IS), outperforming the alignment alternative in generation quality. This demonstrates that preserving ViT encoder's original feature granularity during alignment is beneficial for alignment.

\noindent\textbf{Feature-map upscale choices.} Among the three upscaling options mentioned in Sec.~\ref{sec:experiments}, we figure out that upscaling U-Net hidden states after getting passed through MLP is the best option, achieving 5.72 FID and 161.6 IS. This option is also the most optimal one in terms of computation cost analysis.

\noindent\textbf{Manifold loss weight $w$.} The ablation study on alignment weight $w$ in Eq.~\ref{eq:all_loss} demonstrates a clear performance peak at $w=3$ achieving the lowest FID (5.72) and highest IS (161.6) among tested configurations. 

\subsection{Higher Resolution Experiments}

At the higher-resolution ImageNet 512$\times$512 (w/ cfg) setting, U-REPA remains clearly superior to the REPA baseline (Tab.~\ref{tab:512experiments}). Using SiT$\downarrow$-XL/2, U-REPA reduces FID from 2.44 to 2.21 and raises IS from 247.3 to 274.7. These results indicate that U-REPA's benefits persist at 512 resolution, yielding better distributional fidelity and sample quality/diversity, and demonstrating strong scalability.

\begin{table}[htbp]
  \centering
\begin{tabular}{lcc}
  \toprule
  \multicolumn{3}{l}{\bf{ImageNet} 512$\times$512, w/ cfg} \\
  \toprule
  Alignment Choices & FID$\downarrow$ & IS$\uparrow$ \\
  \midrule
  SiT-XL/2 + REPA & 2.44 & 247.3 \\
  \rowcolor{gray!20}\textbf{SiT$\downarrow$-XL/2 + U-REPA (Ours)} & \textbf{2.21} & \textbf{274.7} \\
  \bottomrule
  \end{tabular}
  \vspace{5pt}
  \caption{\textbf{Comparing U-REPA against REPA on ImageNet 512$\times$512.} On higher resolution, the proposed U-REPA still maintain a clear advantage.}
  \label{tab:512experiments}
\end{table}

\subsection{The Energy Cost Advantage of U-REPA}

We also assess the energy-cost advantage of U-REPA over REPA. We train on eight NVIDIA A100 GPUs and record each GPU's power draw. Combining the measured power with the training duration, we estimate the total energy consumed. The statistics for average power and estimated total energy used by all 8 GPUs are summarized in Tab.~\ref{tab:energy}. Results indicate that our U-REPA method is "greener", costing far less energy.

Reducing training energy directly curbs operational $CO_2$ emissions. Methods that achieve comparable accuracy with lower energy, such as U-REPA vs. REPA in our study, advance both sustainability and the economic viability of large-scale AI.

\begin{table}
\centering
\begin{tabular}{lcccc}
  \toprule
  \multicolumn{5}{l}{\bf{ImageNet} 256$\times$256} \\
  \toprule
  Model & Avg. Pow. (W) & Training Hours & Est. Energy (J) & FID$\downarrow$\\
  \midrule
  \textbf{SiT-XL/2+REPA} (4M iter) & 373.2 & 302.7 & 3.25$\times$10e9 & 1.42 \\
  % \midrule
  \rowcolor{gray!20}\textbf{SiT$\downarrow$-XL/2+U-REPA} (2M iter) & \textbf{295.3} & \textbf{230.7} & \textbf{1.96$\times$10e9} & \textbf{1.41} \\
  \bottomrule
  \end{tabular}
  % }
  \vspace{5pt}
  \caption{\textbf{Energy cost comparison.} We compare the energy cost of U-REPA (at 2M iters) and REPA (at 4M iters). U-REPA could significantly reduce the cost of training a State-of-the-Art diffusion model.}
  \label{tab:energy}
\end{table}

\section{Conclusion}
% \vspace{-5pt}
In this paper, we propose U-REPA, an adapted version of REPA on Diffusion U-Net. We identify key challenges in U-Net hidden state alignment and show that U-REPA effectively bridges the gap between U-Net-based diffusion models and ViT-based encoders. By aligning intermediate features, resolving spatial mismatches via post-MLP upsampling, and enforcing manifold-aware regularization, U-REPA achieves faster convergence and an FID score of 1.41 on ImageNet-256$\times$256 at 2M iters.

\textbf{Acknowledgement.} This work is supported by the National Key R\&D Program of China under grant No. 2022ZD0160300 and the National Natural Science Foundation of China under grant No. 62276007. This work is funded by Peking University--BHP Carbon and Climate Wei-Ming PhD Scholars Program (Program Name: Research on Low-Carbon and Energy-Efficient Large Model Architectures; Program Number: WM202505). We sincerely thank Sibo Fang for his generous help during this project.

\small
\bibliographystyle{plain}
\bibliography{neurips_2025}

%%%%%%%%%%%%%%%%%%%%%%%%%%%%%%%%%%%%%%%%%%%%%%%%%%%%%%%%%%%%

% \iffalse
\newpage
\section*{NeurIPS Paper Checklist}

%%% BEGIN INSTRUCTIONS %%%
The checklist is designed to encourage best practices for responsible machine learning research, addressing issues of reproducibility, transparency, research ethics, and societal impact. Do not remove the checklist: {\bf The papers not including the checklist will be desk rejected.} The checklist should follow the references and precede the (optional) supplemental material.  The checklist does NOT count towards the page
limit. 

Please read the checklist guidelines carefully for information on how to answer these questions. For each question in the checklist:
\begin{itemize}
    \item You should answer \answerYes{}, \answerNo{}, or \answerNA{}.
    \item \answerNA{} means either that the question is Not Applicable for that particular paper or the relevant information is Not Available.
    \item Please provide a short (1–2 sentence) justification right after your answer (even for NA). 
   % \item {\bf The papers not including the checklist will be desk rejected.}
\end{itemize}

{\bf The checklist answers are an integral part of your paper submission.} They are visible to the reviewers, area chairs, senior area chairs, and ethics reviewers. You will be asked to also include it (after eventual revisions) with the final version of your paper, and its final version will be published with the paper.

The reviewers of your paper will be asked to use the checklist as one of the factors in their evaluation. While "\answerYes{}" is generally preferable to "\answerNo{}", it is perfectly acceptable to answer "\answerNo{}" provided a proper justification is given (e.g., "error bars are not reported because it would be too computationally expensive" or "we were unable to find the license for the dataset we used"). In general, answering "\answerNo{}" or "\answerNA{}" is not grounds for rejection. While the questions are phrased in a binary way, we acknowledge that the true answer is often more nuanced, so please just use your best judgment and write a justification to elaborate. All supporting evidence can appear either in the main paper or the supplemental material, provided in appendix. If you answer \answerYes{} to a question, in the justification please point to the section(s) where related material for the question can be found.

IMPORTANT, please:
\begin{itemize}
    \item {\bf Delete this instruction block, but keep the section heading ``NeurIPS paper checklist"},
    \item  {\bf Keep the checklist subsection headings, questions/answers and guidelines below.}
    \item {\bf Do not modify the questions and only use the provided macros for your answers}.
\end{itemize}

%%% END INSTRUCTIONS %%%

\begin{enumerate}

\item {\bf Claims}
    \item[] Question: Do the main claims made in the abstract and introduction accurately reflect the paper's contributions and scope?
    \item[] Answer: \answerYes{} % Replace by \answerYes{}, \answerNo{}, or \answerNA{}.
    \item[] Justification: The abstract demonstrates our motivation, the proposed ideas and a brief summary of experiment results. 

\item {\bf Limitations}
    \item[] Question: Does the paper discuss the limitations of the work performed by the authors?
    \item[] Answer: \answerYes{} % Replace by \answerYes{}, \answerNo{}, or \answerNA{}.
    \item[] Justification: The paper has discussed the limitations of the work in the appendix due to page limits. 

\item {\bf Theory Assumptions and Proofs}
    \item[] Question: For each theoretical result, does the paper provide the full set of assumptions and a complete (and correct) proof?
    \item[] Answer: \answerNA{} % Replace by \answerYes{}, \answerNo{}, or \answerNA{}.
    \item[] Justification: The paper does not inlcude theoretical results.

    \item {\bf Experimental Result Reproducibility}
    \item[] Question: Does the paper fully disclose all the information needed to reproduce the main experimental results of the paper to the extent that it affects the main claims and/or conclusions of the paper (regardless of whether the code and data are provided or not)?
    \item[] Answer: \answerYes{} % Replace by \answerYes{}, \answerNo{}, or \answerNA{}.
    \item[] Justification: The paper fully discloses all the information needed to reproduce the main experimental results of the paper. 

\item {\bf Open access to data and code}
    \item[] Question: Does the paper provide open access to the data and code, with sufficient instructions to faithfully reproduce the main experimental results, as described in supplemental material?
    \item[] Answer: \answerYes{} % Replace by \answerYes{}, \answerNo{}, or \answerNA{}.
    \item[] Justification: The paper will provide open access to the data and code during camera ready period.

\item {\bf Experimental Setting/Details}
    \item[] Question: Does the paper specify all the training and test details (e.g., data splits, hyperparameters, how they were chosen, type of optimizer, etc.) necessary to understand the results?
    \item[] Answer: \answerYes{} % Replace by \answerYes{}, \answerNo{}, or \answerNA{}.
    \item[] Justification: This paper has specified all the training and test details.

\item {\bf Experiment Statistical Significance}
    \item[] Question: Does the paper report error bars suitably and correctly defined or other appropriate information about the statistical significance of the experiments?
    \item[] Answer: \answerNA{} % Replace by \answerYes{}, \answerNo{}, or \answerNA{}.
    \item[] Justification: This is not relevant to this paper.

\item {\bf Experiments Compute Resources}
    \item[] Question: For each experiment, does the paper provide sufficient information on the computer resources (type of compute workers, memory, time of execution) needed to reproduce the experiments?
    \item[] Answer: \answerYes{} % Replace by \answerYes{}, \answerNo{}, or \answerNA{}.
    \item[] Justification: The paper has indicated sufficient information on the computer resources.
    
\item {\bf Code Of Ethics}
    \item[] Question: Does the research conducted in the paper conform, in every respect, with the NeurIPS Code of Ethics \url{https://neurips.cc/public/EthicsGuidelines}?
    \item[] Answer: \answerYes{} % Replace by \answerYes{}, \answerNo{}, or \answerNA{}.
    \item[] Justification: This research conducted in the paper conform, in every respect, with the NeurIPS Code of Ethics.

\item {\bf Broader Impacts}
    \item[] Question: Does the paper discuss both potential positive societal impacts and negative societal impacts of the work performed?
    \item[] Answer: \answerYes{} % Replace by \answerYes{}, \answerNo{}, or \answerNA{}.
    \item[] Discussed in the appendix due to page limits.
    
\item {\bf Safeguards}
    \item[] Question: Does the paper describe safeguards that have been put in place for responsible release of data or models that have a high risk for misuse (e.g., pretrained language models, image generators, or scraped datasets)?
    \item[] Answer: \answerYes{} % Replace by \answerYes{}, \answerNo{}, or \answerNA{}.
    \item[] Justification: This paper has described safeguards.

\item {\bf Licenses for existing assets}
    \item[] Question: Are the creators or original owners of assets (e.g., code, data, models), used in the paper, properly credited and are the license and terms of use explicitly mentioned and properly respected?
    \item[] Answer: \answerYes{} % Replace by \answerYes{}, \answerNo{}, or \answerNA{}.
    \item[] Justification: The utilization of code, data and models in this paper is in accordance with the license and the terms.

\item {\bf New Assets}
    \item[] Question: Are new assets introduced in the paper well documented and is the documentation provided alongside the assets?
    \item[] Answer: \answerNA{} % Replace by \answerYes{}, \answerNo{}, or \answerNA{}.
    \item[] Justification: This paper does not release new assets.

\item {\bf Crowdsourcing and Research with Human Subjects}
    \item[] Question: For crowdsourcing experiments and research with human subjects, does the paper include the full text of instructions given to participants and screenshots, if applicable, as well as details about compensation (if any)? 
    \item[] Answer: \answerNA{} % Replace by \answerYes{}, \answerNo{}, or \answerNA{}.
    \item[] Justification: This paper does not involve crowdsourcing nor research with human subjects.

\item {\bf Institutional Review Board (IRB) Approvals or Equivalent for Research with Human Subjects}
    \item[] Question: Does the paper describe potential risks incurred by study participants, whether such risks were disclosed to the subjects, and whether Institutional Review Board (IRB) approvals (or an equivalent approval/review based on the requirements of your country or institution) were obtained?
    \item[] Answer: \answerNA{} % Replace by \answerYes{}, \answerNo{}, or \answerNA{}.
    \item[] Justification: This paper does not involve crowdsourcing nor research with human subjects.

    \item {\bf Declaration of LLM usage}
    \item[] Question: Does the paper describe the usage of LLMs if it is an important, original, or non-standard component of the core methods in this research? Note that if the LLM is used only for writing, editing, or formatting purposes and does not impact the core methodology, scientific rigorousness, or originality of the research, declaration is not required.
    %this research? 
    \item[] Answer: \answerYes{} % Replace by \answerYes{}, \answerNo{}, or \answerNA{}.
    \item[] Justification: This paper is centered around LLM architecture.

\end{enumerate}
% \fi

% \iffalse
\newpage
\appendix

\section*{Appendix}
\section{Other Improvements}

In the appendix, we address the effect of proposed "Other Improvements" (Sec.~\ref{sec:experiments}), including "Time-aware MLP" and "Weight schedule of loss". 

\noindent\textbf{Time-aware MLP.} Some works on in the diffusion task~\cite{dtr,dynamicdit} reveals that a channel dimension is particularly sensitive and useful to a certain subset of time during sampling. As the time-variant feature is to be aligned to time-invariant vision encoder features, we hold that alignment could perform better when the MLP is time-aware and extract time-invariant information out of diffusion features for alignment.

Inspired by the conditioning of canonical Diffusion U-Net~\cite{ddpm,songunet,dhariwal} and DiT~\cite{dit}, we add a module to predict a pair of channel-wise shift\& scale vector $(\gamma\left(t\right), \beta\left(t\right))$. The module is in parallel to MLP and follows the design of DiT's AdaLN, which is a concatenation of a SiLU and a Linear layer. The shift\& scale vectors are imposed on the output MLP as follows:

$$h^t_\phi(\mathbf{h}_t^{[n]}) = \gamma\left(t\right) \odot h_\phi(\mathbf{h}_t^{[n]}) + \beta\left(t\right) $$

\noindent\textbf{Weight schedule of loss.} As is prompted in REPA~\cite{repa}, designing weight schedule is a future direction. We try various weight schedules (\textit{i.e.} make $\lambda$ in Eq.~\ref{eq:combined_loss} a function $\lambda(t)$ with respect to time) but found that these schedules bring very limited improvements on the proposed U-REPA. Hence, we stick to the original constant weight strategy of REPA.

These improvements bring slight increases on the generation metrics. Hence, we do not include them in the main experiments for the simplicity of the method. The effects of proposed measures are shown in Tab.~\ref{tab:timeaware} and Tab.~\ref{tab:lossschedule}.

\begin{table}[htbp]
  \centering
  \setlength{\belowcaptionskip}{0cm}   
  % \resizebox{\textwidth}{!}{%
\begin{tabular}{lcc}
  \toprule
  \multicolumn{3}{l}{\bf{ImageNet} 256$\times$256, w/ cfg} \\
  \toprule
  Alignment Choices & FID$\downarrow$ & IS$\uparrow$ \\
  \midrule
  \textbf{Ordinary MLP} & {5.72} & {161.6} \\
  \rowcolor{gray!20}\textbf{Time-aware MLP} & \textbf{5.63} & \textbf{163.3} \\
  \bottomrule
  \end{tabular}
  % }
  \vspace{5pt}
  \caption{\textbf{The effect of time-aware MLPs.}}
  \label{tab:timeaware}
  \vspace{-5pt}
\end{table}

\begin{table}[htbp]
  \centering
  \setlength{\belowcaptionskip}{0cm}   
  % \resizebox{\textwidth}{!}{%
\begin{tabular}{lcc}
  \toprule
  \multicolumn{3}{l}{\bf{ImageNet} 256$\times$256, w/ cfg} \\
  \toprule
  Alignment Choices & FID$\downarrow$ & IS$\uparrow$ \\
  \midrule
  \textbf{Constant} & {5.72} & {161.6} \\
  \textbf{$\mathbf{max}(1,t+0.5)$} & {5.85} & {161.3} \\
  \textbf{$\mathbf{max}(1,-t+1.5)$} & {5.72} & {161.6} \\
  \rowcolor{gray!20}\textbf{$\mathbf{min}\left(1, \mathbf{max}\left(-2t+1.5,2t-0.5\right)\right)$} & \textbf{5.58} & \textbf{164.0} \\
  \bottomrule
  \end{tabular}
  % }
  \vspace{5pt}
  \caption{\textbf{The effect of different weight schedules of loss.}}
  \label{tab:lossschedule}
  \vspace{5pt}
\end{table}

\begin{figure*}[htbp]
  \centering
  \includegraphics[width=\textwidth]{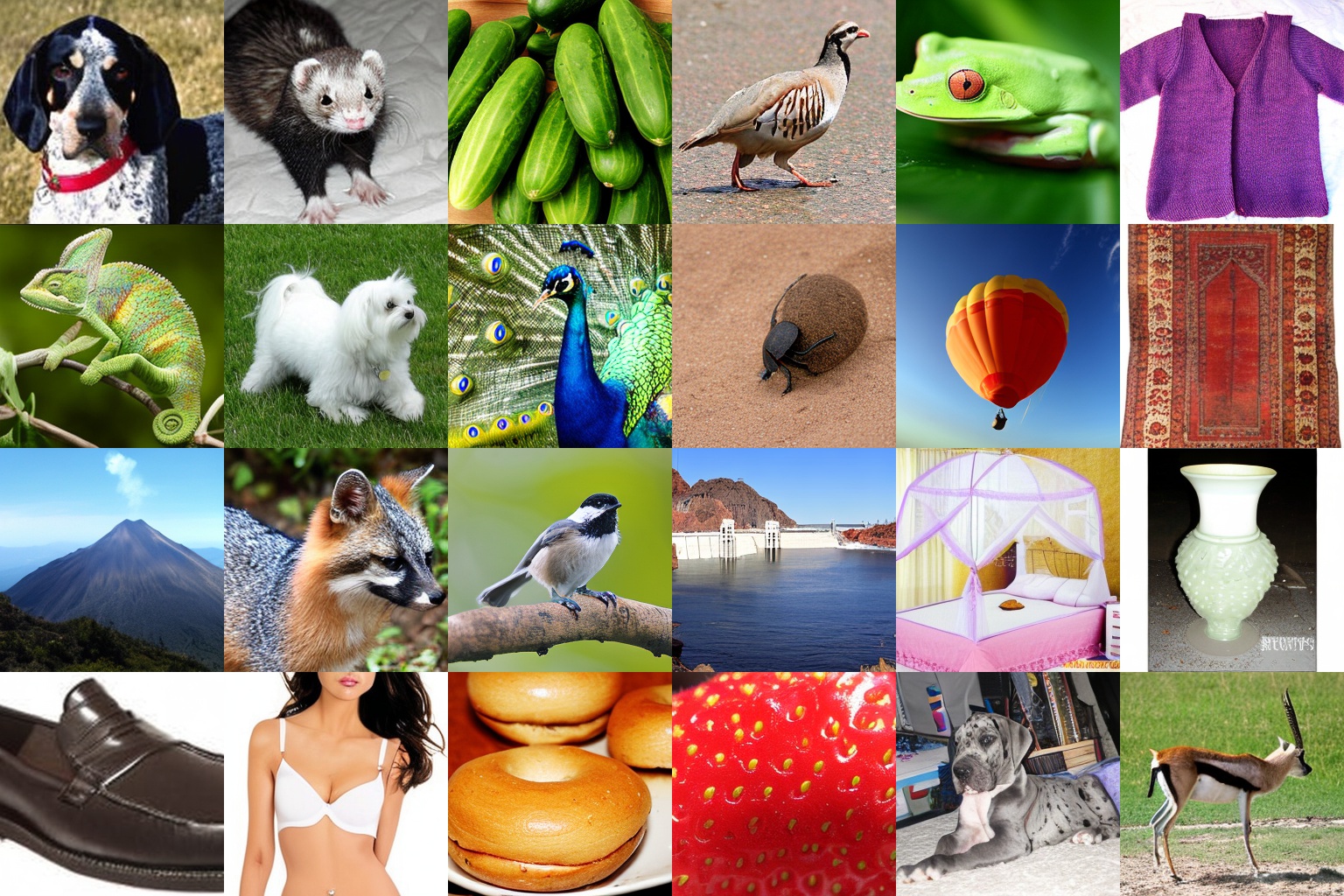}
  \vspace{-10pt}
  \caption{\textbf{Samples generated by SiT$\downarrow$+U-REPA at 1M iterations.} The samples are generated following the setting of REPA, at $cfg=4$. Best viewed on screen.}
  \label{fig:ipt}
  \vspace{-1pt}
\end{figure*}

\section{Additional Experiments}

\textbf{Evaluating SiT$\downarrow$-XL/2.} We also evaluated the proposed U-Net architecture on the Scalable Interpolant Transformers (SiT) framework without the guidance of REPA. The results are shown in Tab.~\ref{tab:sitd}.

Notably, though the amount of FID improvement brought by U-REPA is not as great as REPA (i.e. SiT / SiT+REPA vs. SiT$\downarrow$+U-REPA), we hold that this comparison is invalid due to the following reasons:

\begin{enumerate}
  \item As generation performance gets stronger, it is also becoming much harder to improve (especially for FID when it gets lower).
  \item Aligning SiT$\downarrow$ and ViT is much harder than aligning SiT and ViT, because the backbone of SiT and ViT encoders are very similar. Aligning SiT$\downarrow$ to ViT encoder is a special case due to great architecture difference.
\end{enumerate}

However, we hold that comparing REPA and U-REPA on the same model of SiT$\downarrow$ is fair. The default REPA achieves FID 9.35 (as shown in Tab.~\ref{tab:layer}) while our method achieves FID 5.72, both trained for 100K iterations with cfg and guidance interval adopted.

\begin{table}[htbp]
  \centering
\begin{tabular}{lcc}
  \toprule
  \multicolumn{3}{l}{\bf{ImageNet} 256$\times$256, w/o cfg} \\
  \toprule
  Model & Iter. & FID$\downarrow$ \\
  \midrule
  \textbf{SiT-XL/2} & 400K & 17.2 \\
  \textbf{SiT$\downarrow$-XL/2} & 400K & 9.2 \\
  \textbf{SiT-XL/2+REPA} & 400K & 7.9 \\
  \rowcolor{gray!20}\textbf{SiT$\downarrow$-XL/2+U-REPA} & 400K & \textbf{5.4} \\
  \bottomrule
  \end{tabular}
  % }
  \vspace{5pt}
  \caption{\textbf{Evaluating the performance of SiT$\downarrow$.} SiT$\downarrow$-XL/2 performs much better than SiT-XL/2, and U-REPA further reduces the FID of SiT$\downarrow$-XL/2 to 5.4 without classifier-free guidance.}
  \label{tab:sitd}
\end{table}

\textbf{Seed Sensitivity.} In our paper, we take $seed=0$ following the setting of REPA. We also tested other seeds ($seed=1,2$) in training to examine the seed sensitivity of our method, shown in Tab.~\ref{tab:seed}. The experiments are run for 600K iterations with guidance interval and cfg, following REPA.

\begin{table}[htbp]
  \centering
\begin{tabular}{lcc}
  \toprule
  \multicolumn{3}{l}{\bf{ImageNet} 256$\times$256, w/ cfg} \\
  \toprule
  Model & seed & FID$\downarrow$ \\
  \midrule
  \textbf{SiT$\downarrow$-XL/2+U-REPA} & 0 & 1.618 \\
  \textbf{SiT$\downarrow$-XL/2+U-REPA} & 1 & 1.599 \\
  \textbf{SiT$\downarrow$-XL/2+U-REPA} & 2 & 1.588 \\
  \midrule
  \textbf{SiT$\downarrow$-XL/2+U-REPA} & mean & 1.602$\pm$0.012 \\
  \bottomrule
  \end{tabular}
  % }
  \vspace{5pt}
  \caption{\textbf{Examining seed sensitivity.} We selected $seed=0,1,2$ and evaluate the performance with cfg. The performance fluctuation is limited to a narrow interval (approximately 0.01).}
  \label{tab:seed}
\end{table}

\textbf{Ablations on the REPA Loss.} We also conduct ablations on the REPA loss ($\mathcal{L}_{REPA}$) while leaving the proportion of manifold loss intact (keeping the multiplication $\lambda w$ fixed). The results are shown in Tab.~\ref{tab:lambda_abl}.

\begin{table}[htbp]
  \centering
\begin{tabular}{l>{\columncolor{gray!20}}ccc}
  \toprule
  \multicolumn{4}{l}{\bf{ImageNet} 256$\times$256, w/ cfg} \\
  \toprule
  $\lambda$ & \textbf{0.5} & 0.25 & 0 (No $\mathcal{L}_{REPA}$) \\\midrule
  FID$\downarrow$ & \textbf{5.72} & 6.42 & 10.91 \\
  \bottomrule
  \end{tabular}
  % }
  \vspace{5pt}
  \caption{\textbf{Adjusting the hyperparameter for REPA loss $\lambda$ in Eq.~\ref{eq:all_loss}.} The REPA loss is vital for the generation performance; removing $\mathcal{L}_{REPA}$ would cause a significant performance decay.}
  \label{tab:lambda_abl}
\end{table}

\begin{figure*}[htbp]
  \centering
  \includegraphics[width=\textwidth]{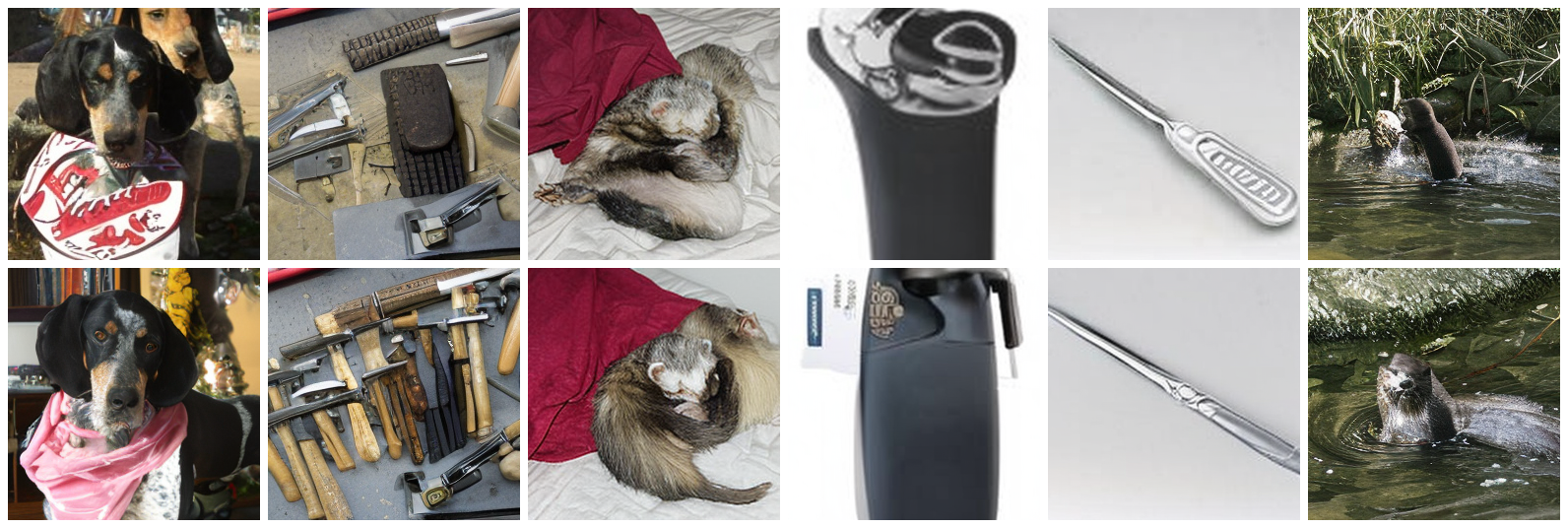}
  \vspace{-10pt}
  \caption{\textbf{Comparing the visual quality of SiT+REPA (upper row) and SiT$\downarrow$+U-REPA (lower row).} The samples are generated following the sampling strategy that yields the State-of-the-Art FIDs in respective methods. Best viewed on screen.}
  \label{fig:1one1}
  \vspace{-1pt}
\end{figure*}

\textbf{One-on-one visualization comparison.} Apart from quantitative comparisons, we also provide qualitative comparisons in Fig.~\ref{fig:1one1} by inserting the same rnadom noise into trained SiT+REPA (at 4M iterations, FID 1.42) and SiT$\downarrow$+U-REPA (at 2M iterations, FID 1.41). The samples are not cherrypicked; we directly pick the first several samples at seed=0. Samples generated by SiT$\downarrow$+U-REPA has better visual quality.

\section{Limitations \& Impact}

\textbf{Limitations and Future work.} The U-Net architecture is a simple one with only one intermediate stage. We do not further refine the architecture as we want to show U-Net architectures as simple as SiT$\downarrow$ could also achieve rapid convergence. Further improvements on the U-Net architecture includes efficient attention~\cite{sana,udit}, non-integer down\& up scaling factors~\cite{fdvit}, and more use of convolutions~\cite{songunet,dic}. Besides, whether U-REPA could be applied to downstream diffusion tasks that rely heavily on U-Nets (e.g. Low-Level Vision) remains to be investigated.

\textbf{Broader Impact.} As a work centered around AIGC, it is probable that inappropriate contents may appear from the output. We should be aware of this negative societal impact.
% Optionally include supplemental material (complete proofs, additional experiments and plots) in appendix.
% All such materials \textbf{SHOULD be included in the main submission.}
% \fi
%%%%%%%%%%%%%%%%%%%%%%%%%%%%%%%%%%%%%%%%%%%%%%%%%%%%%%%%%%%%

\end{document}